\documentclass[runningheads]{llncs}
\usepackage{graphicx}
\usepackage{amsmath,amssymb} % define this before the line numbering.
\usepackage{color}\usepackage[width=122mm,left=12mm,paperwidth=146mm,height=193mm,top=12mm,paperheight=217mm]{geometry}
\usepackage{caption}
\usepackage{subcaption}
\usepackage{float}
\usepackage{authblk}

\usepackage{array}
\newcolumntype{P}[1]{>{\centering\arraybackslash}p{#1}}
\usepackage{booktabs}
\newcommand{\ra}[1]{\renewcommand{\arraystretch}{#1}}

\begin{document}
\pagestyle{headings}
\mainmatter
\title{To Fall Or Not To Fall:\\A Visual Approach to Physical Stability Prediction} % Replace with your title

\titlerunning{A Visual Approach to Physical Stability Prediction}

\authorrunning{Wenbin Li, Seyedmajid Azimi, Ale\v{s} Leonardis, Mario Fritz}

\author{Wenbin Li\textsuperscript{1}, Seyedmajid Azimi\textsuperscript{1}, Ale\v{s} Leonardis\textsuperscript{2}, Mario Fritz\textsuperscript{1}}
\institute{\textsuperscript{1}Max Planck Institute for Informatics, Saarbr{\" u}cken, Germany\\ \textsuperscript{2}School of Computer Science, University of Birmingham, United Kingdom}

\maketitle

\begin{abstract}
Understanding physical phenomena is a key competence that enables humans and animals to act and interact under uncertain perception in previously unseen environments containing novel object and their configurations. Developmental psychology has shown that such skills are acquired by infants from observations at a very early stage.

In this paper, we contrast a more traditional approach of taking a model-based route with explicit 3D representations and physical simulation by an {\em end-to-end} approach that directly predicts stability and related quantities from appearance. We ask the question if and to what extent and quality such a skill can directly be acquired in a data-driven way---bypassing the need for an explicit simulation.

We present a learning-based approach based on simulated data that predicts stability of towers comprised of wooden blocks under different conditions and quantities related to the potential fall of the towers. The evaluation is carried out on synthetic data and compared to human judgments on the same stimuli.

\end{abstract}

\section{Introduction}
Scene understanding requires -- among others -- understanding of relations between and among the objects. Many of these relations are governed by the Newtonian laws and thereby rule out unlikely or even implausible configurations for the observer. They are part of ``dark matter'' \cite{Xie_2013_ICCV} in our everyday visual data which helps us {\em interpret} the configurations of  objects correctly and accurately. 
Although objects simply obey these elementary laws of Newtonian mechanics, which can very well be captured in simulators, uncertainty in perception makes exploiting these relations challenging in artificial systems.

In contrast, humans understand such physical relations naturally, which e.g. enables them to manipulate and interact with novel objects in unseen conditions with ease. We build on a rich set of prior experiences that allow us to employ a type of commonsense understanding that does not -- most likely -- involve symbolic representation of 3D geometry that is processed by a physics simulation engine. We rather seem to build on what has been coined as ``na\"{i}ve physics'' \cite{Smith1994} or ``intuitive physics'' \cite{mccloskey1983intuitive}, 
which is a good enough proxy to make us operate successfully in the real-world.

It has not been shown yet how to equip machines with a similar set of physics commonsense -- and thereby bypassing a model--based representation and a physical simulation. In fact, it has been argued that such an approach is unlikely due to e.g. the complexity of the problem \cite{battaglia2013simulation}. Only recently, several approach have revived this idea and reattempted a fully data drive approach to capturing the essence of physical events via machine learning methods \cite{mottaghi2015newtonian,wu2015galileo,fragkiadaki2015learning}.

\begin{figure}
\centering
        \begin{subfigure}[b]{0.26\textwidth}
                \centering
                \includegraphics[width=.85\linewidth]{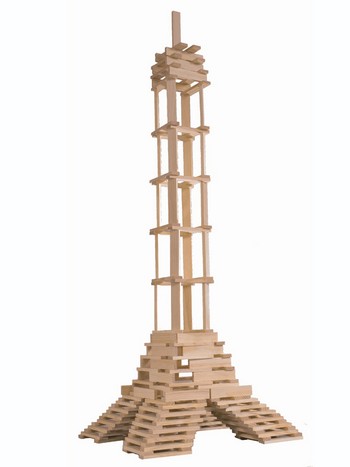}
                \caption{Kapla Blocks}
                \label{fig:kapla}
        \end{subfigure}%
        \begin{subfigure}[b]{0.26\textwidth}
                \centering
                \includegraphics[width=.85\linewidth]{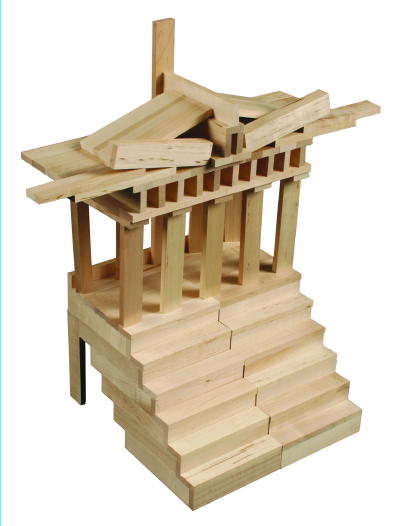}
                \caption{KEVA planks}
                \label{fig:keva}
        \end{subfigure}%
        \begin{subfigure}[b]{0.26\textwidth}
                \centering
                \includegraphics[width=.85\linewidth]{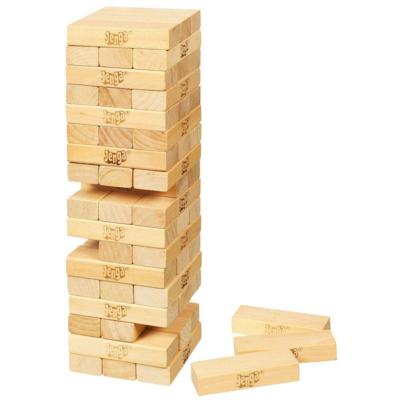}
                \caption{Jenga}
                \label{fig:jenga}
        \end{subfigure}%
        \caption{Examples toys embody the support event.}
        \label{fig:toys}
\end{figure}

In contrast, studies in developmental psychology \cite{baillargeon1995model} have shown that infants acquire the knowledge of physical events by observation at a very early age, including support, collision and unveiling. 
According to their research, the infant with some innate basic core knowledge \cite{baillargeon2008innate} gradually builds its internal model of the physical event by observing its various outcomes. Amazingly, such basic knowledge of physical event, for example the understanding of support phenomenon can make its way into relatively complex operations as shown in Figure~\ref{fig:toys}. Such structures are generated by stacking up an element or removing one while retaining the structure's stability primarily relying on effective knowledge of support events in such toy constructions. In our work, we focus on exactly this support event and construct a model for machines to predict object stability.

Hence, we revisit the classic setup of Tenenbaum and colleagues \cite{battaglia2013simulation} and explore to which extend machines can predict physical stability events directly from appearance cues. 
We approach this problem by synthetically generating a large set of wood block towers with a range of conditions, including varying number of blocks, varying block sizes, more planar vs. multi-layered configurations. We run those configurations through a simulator ({\em only at training time!} ) in order to generate labels if the tower would fall. We show for the first time that aforementioned stability test can be learned and predicted in a purely data driven way -- bypassing traditional model-based simulation approaches. In order to shed more light on the capabilities and limitations of our model, we accompany our experimental study with human judgments on the same stimuli.
\section{Related Work}
As human, we possesses the ability to judge from vision alone if an object is physically stable or not and predict the objects' physical behaviors. Yet it is unclear: (1) how do we
make such decision and (2) how do we acquire this capability. Research in development psychology \cite{baillargeon1994infants,baillargeon1995model,baillargeon2002acquisition} suggests that infants acquire the knowledge of physical events at very young age by observing those events, including support events and others. This partly answers to the question (2), however there seems no consensus on how the internal mechanism for interpreting external physical events to address question (1). \cite{battaglia2013simulation} proposed an intuitive physics simulation engine for such mechanism and found it resemble to human subjects' behaviors pattern in several psychological tasks. 
Historically,  intuitive physics is connected to the case where people often hold erroneous physical intuitions \cite{mccloskey1983intuitive}, such as people tend to expect an object dropped from a moving subject will fall vertically straight down. It is rather counter-intuitive how the proposed simulation engine in \cite{battaglia2013simulation} can explain such erroneous intuitions. 

While it is probably illusive to fully reveal human's inner mechanism for physical modeling and inference, it is feasible to build up models based on observation, in particular the visual information. In fact, looking back to history, physical laws are discovered through the  observation of physical events \cite{macdougal2012galileo}. Our work is in this direction. By observing a large number of support event instances in simulation, we want to gain deeper insight into the prediction paradigm.  

In our work, we use a game engine to render scene images and a built-in physics simulator to simulate the scenes' stability behavior. The data generation procedure is based on the platform used in \cite{battaglia2013simulation}, however as discussed before, their work hypothesized a simulation engine as an internal mechanism for human to understand the physics in the external world while we are interested in finding an image-based model to directly predict the physical behavior from visual channel. Learning from synthetic data has a long tradition in computer vision and recently has gained increasing interest \cite{li12eccv,kostas14cvpr,peng2015learning,rematas16cvpr} due to data hungry deep learning approach.

Understanding physical events also plays an important role in scene understanding in computer vision. By including the additional clue from physical constraints into the inference mechanism, mostly from the support event, it has further improved results in segmentation of surfaces \cite{gupta2010blocks}, scenes \cite{silberman2012indoor} from image data, and object segmentation in 3D point cloud data \cite{zheng2013beyond}.

Only very recently, learning physical concepts from data has been attempted.
\cite{mottaghi2015newtonian} aims at understanding dynamic events governed by laws of Newtonian physics, but uses proto-typical motion scenarios as exemplars.
\cite{fragkiadaki2015learning} analyze a billiard table scenarios and aim at learning the dynamics from observation. While learning is largely data-driven, the object notion is predefined as location of the balls are provided to the system.
\cite{wu2015galileo} aims to understand physical properties of objects. They again rely on a explicit physical simulation.

In contrast, we only use simulation at training time and predict for the first time visual stability directly from visual inputs of towers with a large number of degrees of freedom.

A recent paper \cite{fergus16blocsarxiv} that appeared a few days before this work is a related research thread that was conducted in parallel without our knowledge. The focus of their work is different from ours, namely predicting outcome and falling trajectories for simple 4 block scenes. 
In our work, we significantly vary the scene parameters, investigate if and how the prediction performance from image trained model changes according to such changes, and further we examine how the human's prediction adapt to the variation in the generated scenes and compare it to our model.
\section{Towards Modeling a Visual Stability Test}
In order to tackle a visual stability test, we require a data generation process that allows us to control various degrees of freedom induced by the problem as well as generation of large quantities of data in a repeatable setup. Therefore, we follow the pioneer work on this topic \cite{battaglia2013simulation} and use a simulator to setup and predict physical outcomes of wood block towers. Afterwards, we describe the method that we investigate for visual stability prediction. We employ state of the art deep learning techniques, which are the de facto standard in today's recognition systems. Lastly, we describe the setup of the human study that we conduct to complement the machine predictions with a human reference.

\subsection{Synthetic Data}
Based on the scene simulation framework used in \cite{hamrick2011internal,battaglia2013simulation}, we generate synthetic data in our experiment with rectangular cuboid blocks as basic elements. Number of blocks, block size, stacking depth are varied in different scenes, to which we will refer as {\it scene parameters}. 
\paragraph{Block Numbers}
We expect that varying size of towers and involved blocks will influence the difficulty and challenge the competence of eye-balling the stability of a tower in humans and machine. While evidently the appearance becomes more complex, with increasing number of blocks, the number of contact surfaces and interactions equally make the problem richer. Therefore, we include scenes with four different numbers of blocks, 4 blocks, 6 blocks, 10 blocks and 14 blocks as $\{4B, 6B, 10B, 14B\}$.

\begin{figure}
\centering
        \begin{subfigure}[b]{0.4\textwidth}
                \centering
                \includegraphics[width=.85\linewidth]{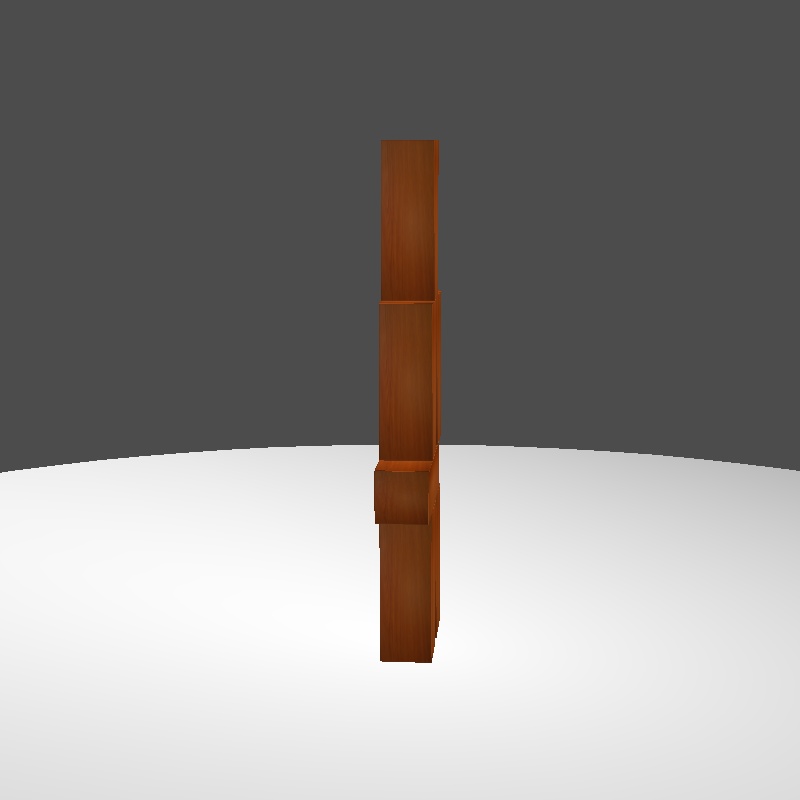}
                \caption{2D stacking}
                \label{fig:2d}
        \end{subfigure}%
        \begin{subfigure}[b]{0.4\textwidth}
                \centering
                \includegraphics[width=.85\linewidth]{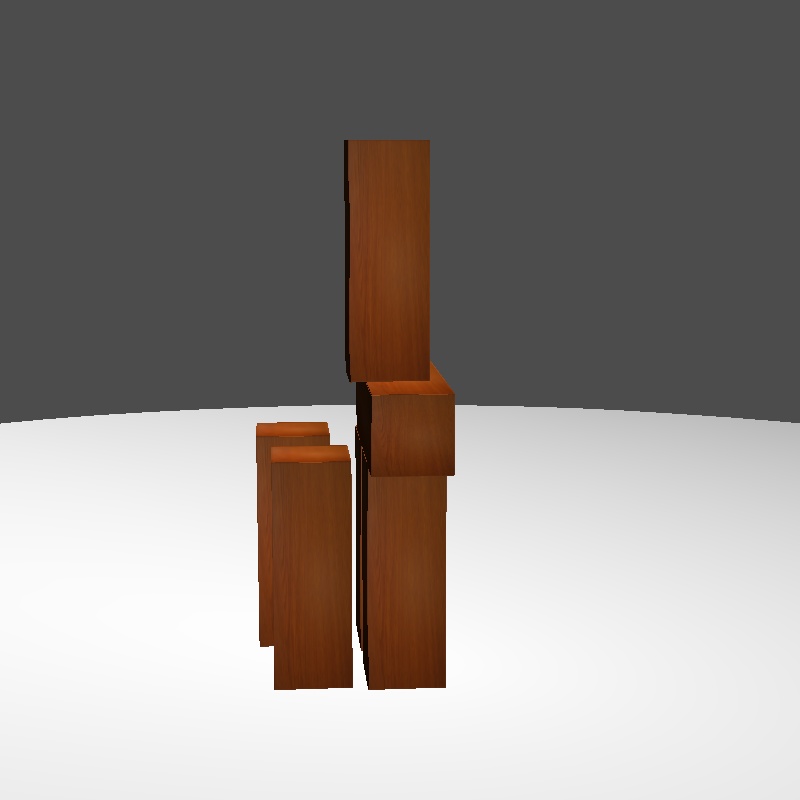}
                \caption{3D stacking}
                \label{fig:3d}
        \end{subfigure}%
        \caption{Example scenes for 2D and 3D stacking with 6 blocks from side view.}
        \label{fig:2dv3d}
\end{figure}

\paragraph{Stacking Depth}
As we focus our investigations on judging stability from monocular input, we vary the depth of the tower from a one layer setting which we call $2D$ to a multi-layer setting which we call $3D$.
The first one only allows a single block along the image plane at all height levels while the other does not enforce such constraint and can expand in the image plane. Visually, the former results in a single-layer stacking similar to Tetris while the latter ends in a multiple-layer structure as shown in Figure~\ref{fig:2dv3d}. The latter most likely requires the observer to pick up on more subtle visual cues, as its layers are heavily occluded. 

\paragraph{Block Size}
We include two groups of block size settings. In the first one, the towers are constructed of blocks that have all the same size of $0.2m\times 0.2m\times 0.6 m$ as in the \cite{battaglia2013simulation}. The second one introduces varying block sizes where two of the three dimensions are randomly scaled with respect to a truncated Normal distribution $N(1,\sigma^2)$ around $[1 - \delta,1 + \delta]$,  $\sigma$ and $\delta$ are small values. These two settings are referred to as $\{Uni, NonUni\}$. The setting with non uniform blocks introduces small visual cues where stability hinges on small gaps between differently sized blocks that are challenging even for human observers.

\paragraph{Scenes}
Combining these three scene parameters, we define $16$ different scene groups. For example, group 10B-2D-Uni is for scenes stacked with 10 Blocks of same size, stacked within a single layer. For each group, $1000$ candidate scenes are generated where each scene is constructed with non-overlapping geometrical constraint in a bottom-up manner. There are $16 K$ scenes in total. For prediction experiments, half of the images in each group are for training and the other half for test, the split is fixed across the experiments.

\paragraph{Rendering}
While we keep the rendering basic, we like to point out that we deliberately decided against colored bricks as in \cite{battaglia2013simulation} in order to challenge perception and make identifying brick outlines and configurations more challenging. The lighting is fixed across scenes and the camera is automatically adjusted so that the whole tower is centered in the captured image. Images are rendered at resolution of $800 \times 800$ in color.

\paragraph{Physics Engine}
We use Bullet \cite{coumans2010bullet} physics engine in Panda3D \cite{goslin2004panda3d} to perform physics-based simulation for $2000 ms$ at $1000Hz$ for each scene. Surface friction and gravity are enabled in the simulation. The system records the configuration of a scene of $N$ blocks at time $t$ as $(p_1,p_2,...,p_N)_t$, where $p_i$ is the location for block $i$. The stability is then automatically decided as a Boolean variable: 
\begin{gather*}
\begin{aligned}
S = \bigwedge\limits_{i=1}^N (\Delta((p_i)_{t=T} - (p_i)_{t=0}) > \tau)
\end{aligned}
\end{gather*}
where $T$ is the end time of simulation, $\delta$ measures the displacement for the blocks between the starting point and end time, $\tau$ is the displacement threshold, $\wedge$ denotes the logical {\it and} operator, that is to say it counts as unstable $S=True$ if any block in the scene moved in simulation, otherwise as stable $S=False$.

\subsection{Stability Prediction from Still Images}
\paragraph{Inspiration from Human}
Research in \cite{hamrick2011internal,battaglia2013simulation} suggests the combinations of the most salient features in the scenes are insufficient to capture people's judgments, however, contemporary study reveals human's perception of visual information, in particular some geometric feature, like critical angle \cite{cholewiak2013visual,cholewiak2015perception} plays an important role in the process. Regardless of the actual inner mechanism for human to parse the visual input, it is clear there is a mapping $f$ involving visual input $I$ to the stability prediction $P$. 
\begin{gather*}
\begin{aligned}
f: I,\ast \rightarrow P
\end{aligned}
\end{gather*}
The mapping can be inclusive, as in \cite{hamrick2011internal} using it along with other aspects, like physical constraint to make judgment or exclusive, as in \cite{cholewiak2013visual} using visual cues alone to decide.

\paragraph{Image Classifier for Stability Prediction}
In our work, we are interested in the mapping $f$ exclusive to visual input and directly predicts the physical stability. We use deep convolutional neural networks to learn the mapping as it has shown great success on image classification task \cite{krizhevsky2012imagenet}. Such networks have been shown to be able to adapt to a wide range of classification and prediction task \cite{razavian2014cnn} through re-training or adaptation by fine-tuning.
Therefore, these approaches seem adequate method to study visual prediction on this challenging task with the motivation that by changing conventional image classes labels to stability labels the network can learn ``physical stability salient'' features.

In a pilot study, we tested on a subset of the generated data with LeNet \cite{lecun1995comparison}, a relatively small network designed for digit recognition, AlexNet \cite{krizhevsky2012imagenet}, a large network and VGG Net\cite{simonyan2014very}, a even larger network than AlexNet. We trained from scratch for the LeNet and fine-tuned for the large network pre-trained on ImageNet \cite{deng2009imagenet}. VGG Net consistently outperforms the other two, hence we use it across our experiment.  We use the Caffe framework \cite{jia2014caffe} in all our experiments.

\subsection{Human Subject Study}
We recruit human subjects to predict stability for give scene images.
Due to large number of test data, we sample images from different scene groups for human subject test. 8 subjects are recruited for the test. Each subject is presented with a set of captured images from the test split. Each set includes $96$ images where images cover all $16$ scene groups with $6$ scene instances per group. For each scene image, subject is required to rate the stability on a scale from $1-5$ without any constraint for response time:
\begin{enumerate}
\item
Definitely unstable: definitely at least one block will move/fall
\item
Probably unstable: probably at least one block will move/fall
\item
Cannot tell: the subject is not sure about the stability
\item
Probably stable: probably no block will move/fall
\item
Definitely stable: definitely no block will move/fall
\end{enumerate}

\noindent
The main questions that we aim at answering with  our study are as follows:
\begin{enumerate}
\item
How well do humans perform on the task?

This provides a basic reference of the task difficulty. High human performance may indicate the task is too simple while low performance for too difficulty for the setup. In addition, human performance is also a baseline to our image-based model.

\item
How does human performance vary with respect to scene variations?

This provides an additional reference to the task difficulty concerning the scene parameter. If certain scene parameter does not influence too much on the human performance, it can be less relevant to human perception for the task. If the human performance changes along with the variation of the scene parameter, then we can further distinguish between more dominated factors and less dominated ones. Moreover, it serves as a parallel baseline to compare with the image-based model to see if both are affected by the presented challenges equally.

\item
How does human performance compare to image-based model?

This answers if the data-driven image-based approach can match or even possible to outperform human.  

\item
How do human vs. machine confidences relate? 

The setup of our human study provides as with confidence values 1-5, that reflect the certainty in the human judgment. We would like to investigate how these confidences are mimicked by our data-drive approach.

\item
How do failure cases differ? Are they plausible?

In order to gain more insights in our model and analyze to what extend a scene understanding of real physics or a certain human notion of commonsense physics  was achieved, we can compare failure cases and modes between human and machine prediction.

\end{enumerate}

Since we want to compare the performance between the image-based model and human, we use the same data for both in test. More details will be discussed in the following section.

\section{Experiment}
Our experimental analysis is composed of two main part. The first will study if and to what extend a largely model free visual stability test can be learned directly from data only. The second part, will put these finding in relation to human judgments on the same stimuli.
%--------------------------------------------------------------------------------------------------------------------------------------
\subsection {Visual Stability Prediction}
In this part of experiments, image are captured before the physics engine is enabled, and the stability labels are recorded from the simulation engine as described before. At training time, the model has access to image and the stability labels. At test time, the learned model predicts stability results against the results generated by the simulator.

We divide the experiment design into 3 sets: the intra-group, cross-group and generalization. The first set investigates influence on the model's performance from individual scene parameter, the other two sets explore generalization properties under different settings.

%--------------------------------------------------------------------------------------------------------------------------------------
\subsubsection{Intra-Group Experiment}
In this set of experiments, we train and test on the scenes with the same scene parameters in order to assess the feasibility of our task.

\paragraph{Number of Blocks}
In this group of experiment, we fix the stacking depth and keep the all blocks in the same size but vary the number of blocks in the scene to observe how it affects the prediction rates from the image trained model, which approximates the relative recognition difficulty from this scene parameter alone. The results have been shown in Table~\ref{tab:intra_group}. Consistent drop of performance can be observed with increasing number of blocks in the scene under various block sizes and stacking depth conditions. More blocks in the scene generally leads to higher scene structure and hence higher difficulty in perception. 

\begin{figure}
\centering
        \begin{subfigure}[b]{0.25\textwidth}
                \centering
                \includegraphics[width=.85\linewidth]{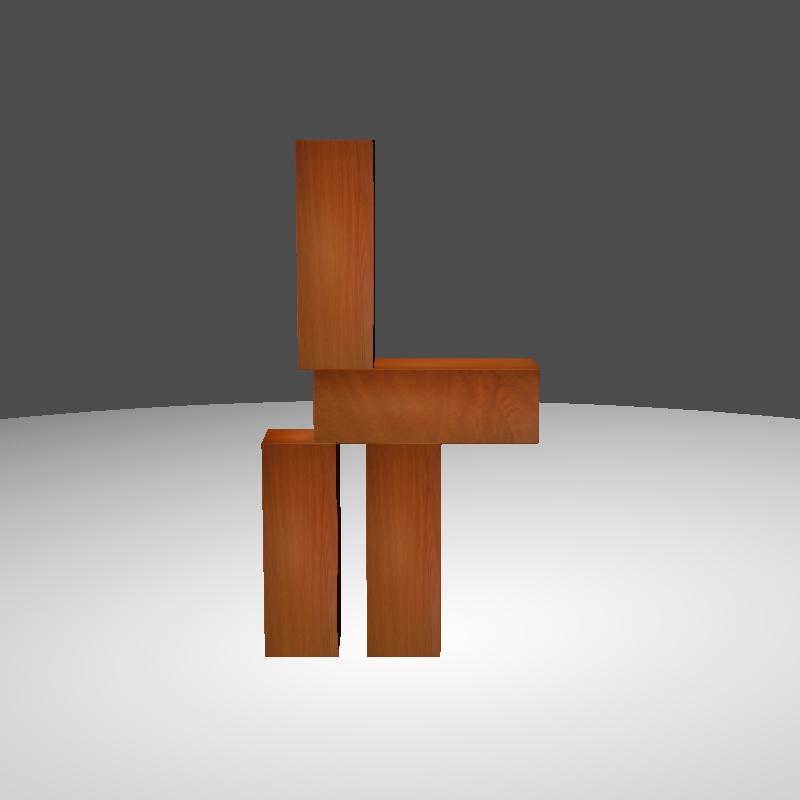}
                \caption{4 Blocks}
                \label{fig:4blks}
        \end{subfigure}%
        \begin{subfigure}[b]{0.25\textwidth}
                \centering
                \includegraphics[width=.85\linewidth]{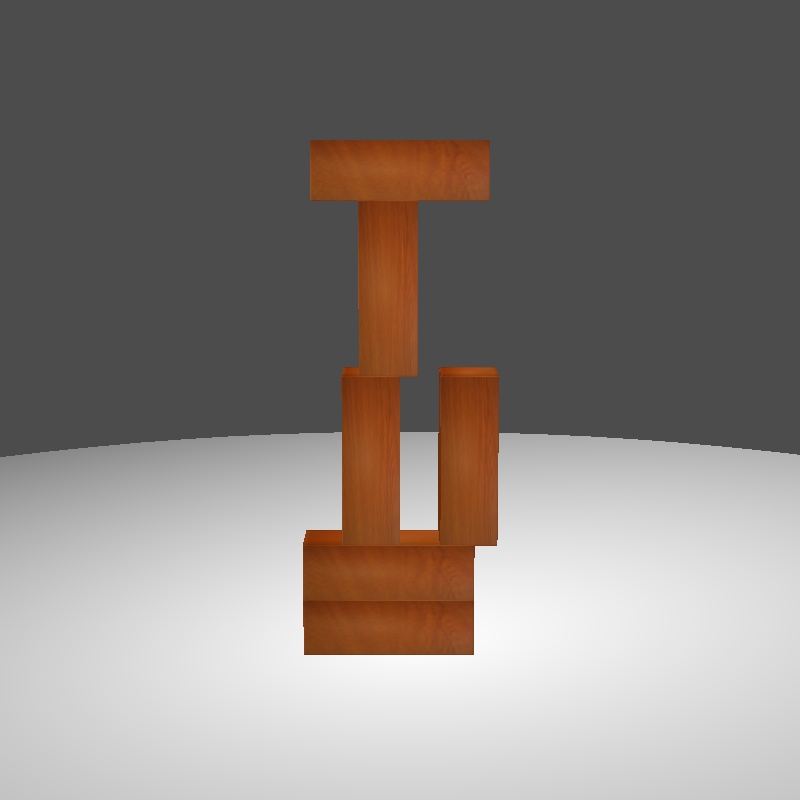}
                \caption{6 Blocks}
                \label{fig:6blks}
        \end{subfigure}%
        \begin{subfigure}[b]{0.25\textwidth}
                \centering
                \includegraphics[width=.85\linewidth]{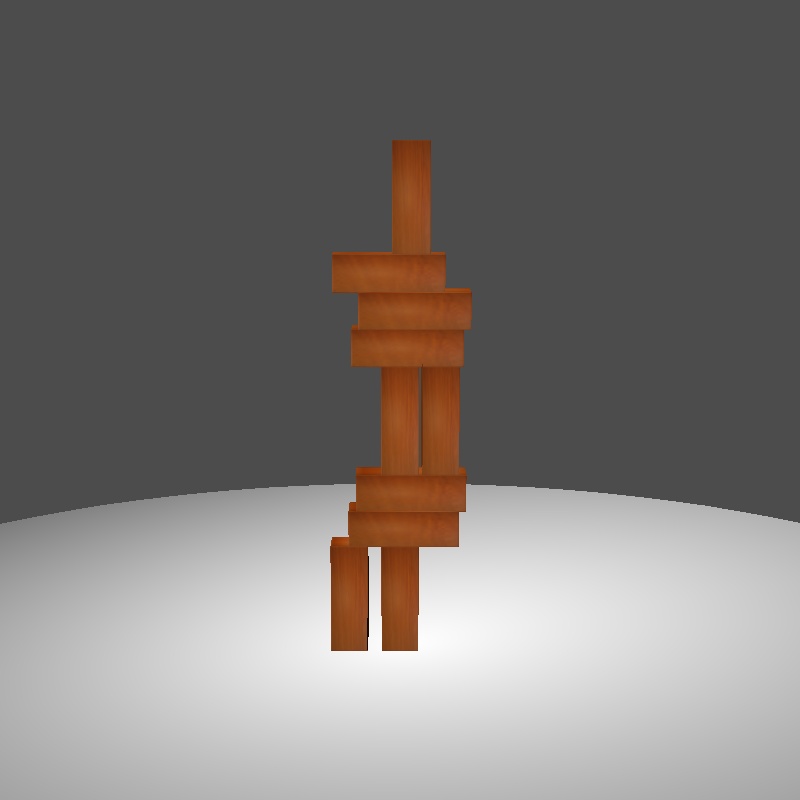}
                \caption{10 Blocks}
                \label{fig:10blks}
        \end{subfigure}%
        \begin{subfigure}[b]{0.25\textwidth}
                \centering
                \includegraphics[width=.85\linewidth]{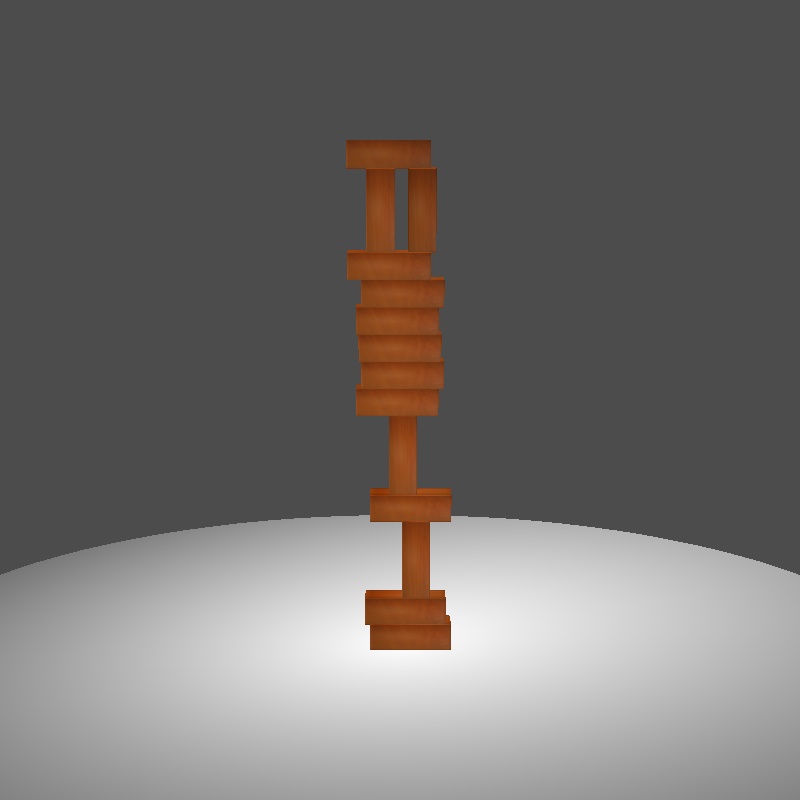}
                \caption{14 Blocks}
                \label{fig:14blks}
        \end{subfigure}
        \caption{Examples of scenes with different number of blocks}
        \label{fig:vary_blks}
\end{figure}

\paragraph{Block Size}
In this group of experiment, we aim to explore how same size and varied blocks sizes affect the prediction rates from the image trained model. We compare the results at different number of blocks to the previous group, in the most obvious case,  scenes happened to have similar stacking patterns and same number of blocks can result in changes visual appearance. To further eliminate the influence from the stacking depth, we fix all the scenes in this group to be 2D stacking only. As can be seen from Table~\ref{tab:intra_group}, the performance decreases when moving from 2D stacking to 3D. The additional variety introduced by the block size indeed makes the task more challenging.

\begin{figure}
\centering
        \begin{subfigure}[b]{0.32\textwidth}
                \centering
                \includegraphics[width=.85\linewidth]{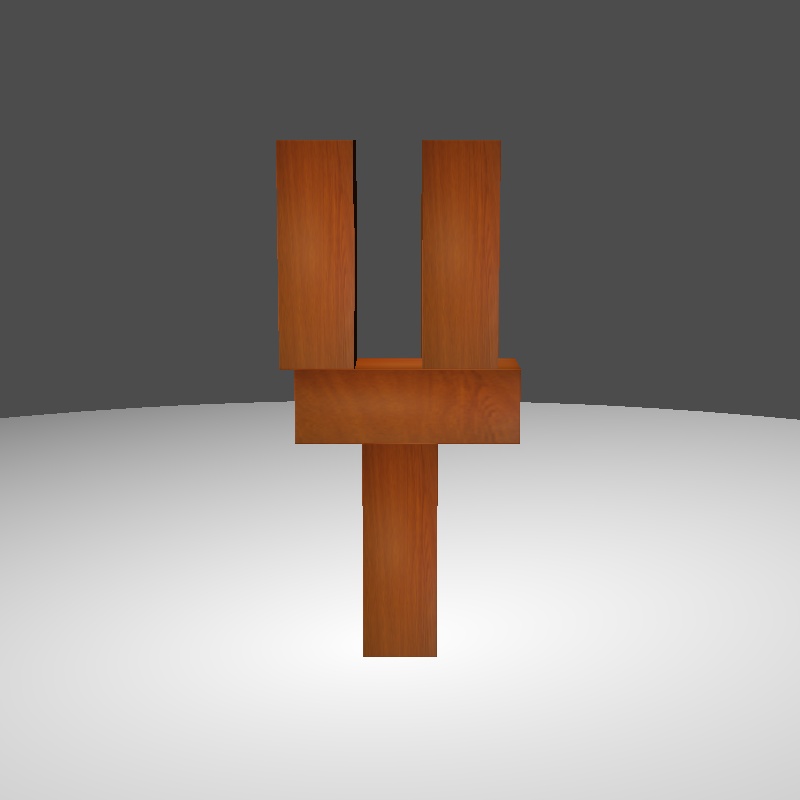}
                \caption{Fixed blocks size}
                \label{fig:4blks}
        \end{subfigure}%
        \begin{subfigure}[b]{0.32\textwidth}
                \centering
                \includegraphics[width=.85\linewidth]{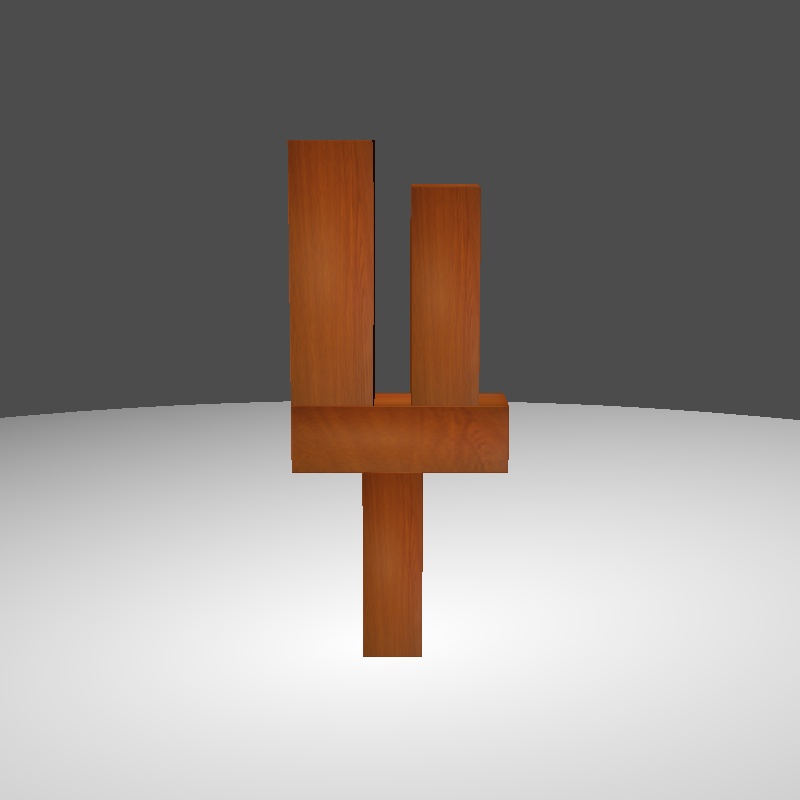}
                \caption{Varied blocks size}
                \label{fig:6blks}
        \end{subfigure}%
        \caption{Examples of scenes with the fixed size and varied sizes.}
        \label{fig:vary_szs}
\end{figure}

\paragraph{Stacking Depth}
In this group of experiment, we want to investigate how stacking depth affects the prediction rates. With increasing stacking depth, it naturally introduces ambiguity in the perception of the scene structure, namely some parts of the scene can be occluded or partially occluded by other parts. Similar to the experiments in previous groups, we want to minimize the influences from other scene parameters, we fix the block size to be the same and only observe the performance across different number of blocks. The results in Table~\ref{tab:intra_group} show a little inconsistent behaviors between relative simple scenes (4 blocks and 6 blocks) and difficult scenes (10 blocks and 14 blocks). For simple scenes, prediction accuracy increases when moving from $2D$ stacking to $3D$ while it is the other way around for the complex scene. Naturally relaxing the constraint in stacking depth can introduce additional challenge for perception of depth information, yet given a fixed number of blocks in the scene, the condition change is also more likely to make the scene structure lower which reduces the difficulty in perception. A combination of these two factors decides the final difficulty of the task, for simple scenes, the height factor has stronger influence and hence exhibits better prediction accuracy for $3D$ over $2D$ stacking while for complex scenes, the stacking depth dominates the influence as the significant higher number of blocks can retain a reasonable height of the structure, hence receives decreased performance when moving from $2D$ stacking to $3D$.

\begin{table*}\setlength{\tabcolsep}{7pt}
\centering
\ra{1.4}
\begin{tabular*}{0.6\textwidth}{@{ }ccccc@{ }}\toprule
Num.of Blks & \multicolumn{2}{c}{Uni.} & \phantom{ab}& \multicolumn{1}{c}{NonUni.}\\
\cmidrule{2-3} \cmidrule{5-5}
 & $2D$ & $3D$ && $2D$\\ \midrule
$4B$ & 93.0 & 99.2 && 93.2\\
$6B$ & 88.8& 91.6&& 88.0\\
$10B$ & 76.4& 68.4&& 69.8\\
$14B$ & 71.2& 57.0&& 74.8\\ \bottomrule
\end{tabular*}
\caption{Intra-group experiment by varying scene parameters.} % from prediction files
\label{tab:intra_group}
\end{table*}

\begin{figure}
\centering
       \begin{subfigure}[b]{0.25\textwidth}
                \centering
                \includegraphics[width=.85\linewidth]{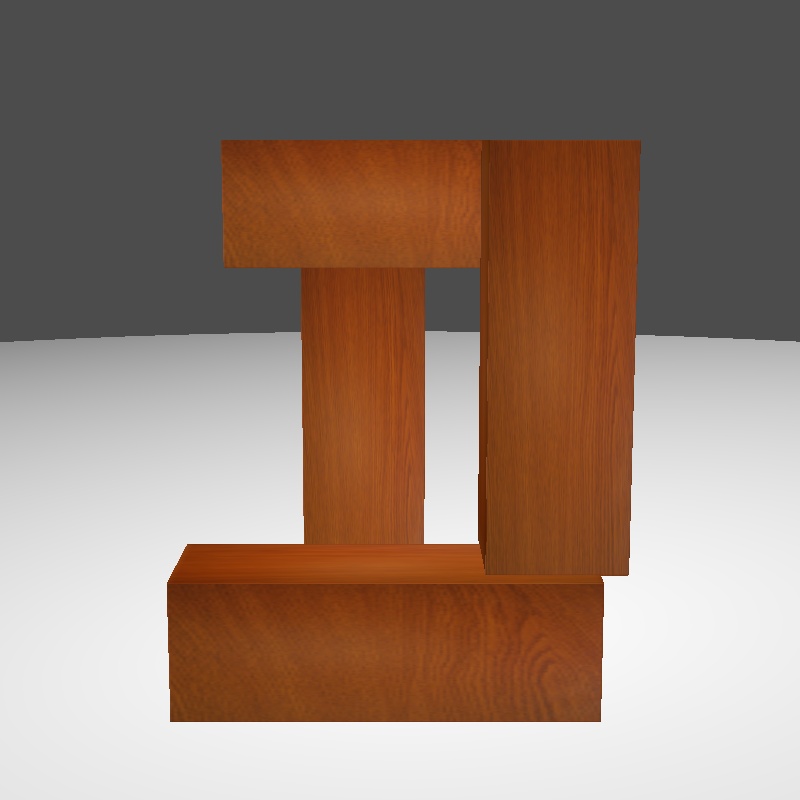}
                \caption{4 blocks 3D}
                \label{fig:4blks}
        \end{subfigure}%
        \begin{subfigure}[b]{0.25\textwidth}
                \centering
                \includegraphics[width=.85\linewidth]{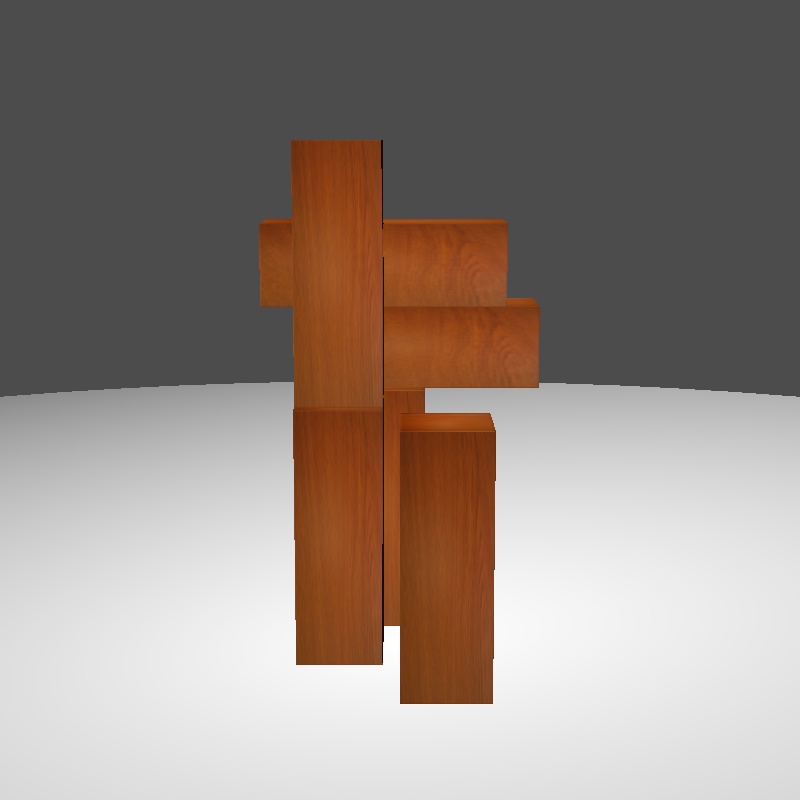}
                \caption{6 blocks 3D}
                \label{fig:4blks}
        \end{subfigure}%
        \begin{subfigure}[b]{0.25\textwidth}
                \centering
                \includegraphics[width=.85\linewidth]{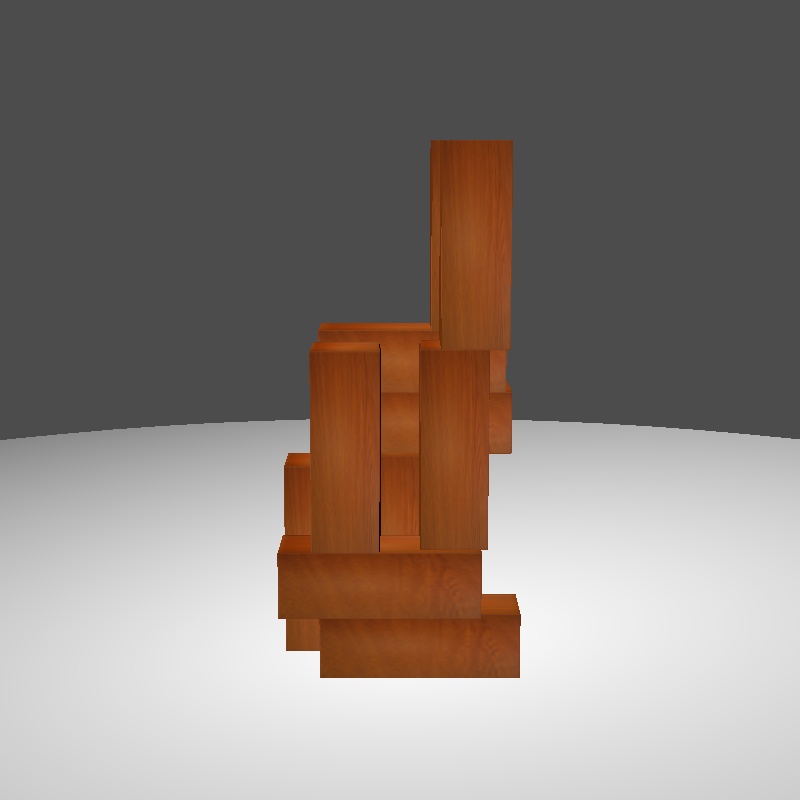}
                \caption{10 blocks 3D}
                \label{fig:6blks}
        \end{subfigure}%
         \begin{subfigure}[b]{0.25\textwidth}
                \centering
                \includegraphics[width=.85\linewidth]{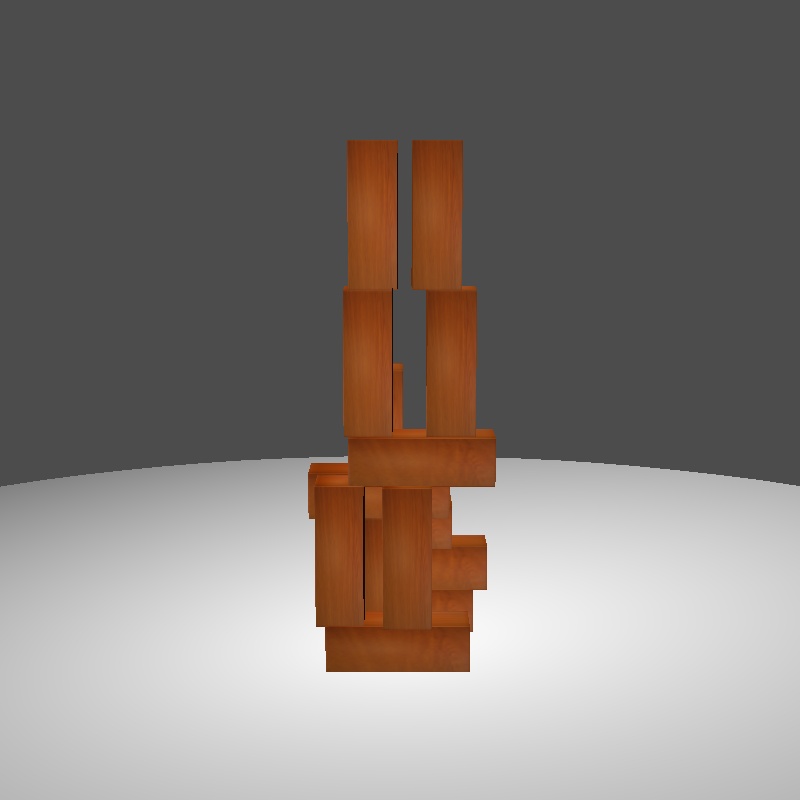}
                \caption{14 blocks 3D}
                \label{fig:6blks}
        \end{subfigure}%
        \caption{Examples of scenes with 3D stacking.}
        \label{fig:vary_dims}
\end{figure}

%--------------------------------------------------------------------------------------------------------------------------------------
\subsubsection{Cross-Group Experiment}
\begin{figure*}
\centering
\includegraphics[width=0.9\linewidth]{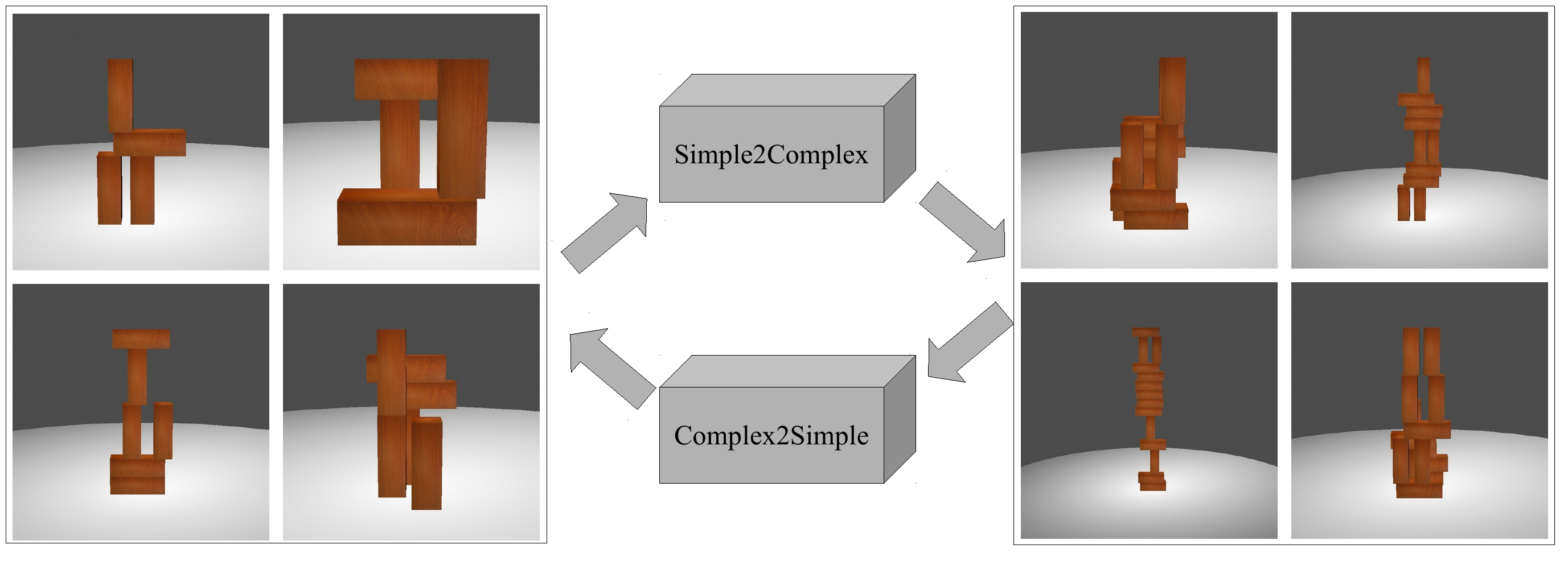}
\caption{Experiment settings for Cross-group classification.}
\label{fig:cross_group}
\end{figure*}

In this set of experiment, we want to see how the learned model transfers across scenes with different complexity, so we further divide the scene groups into two large groups by the number of blocks, where a {\it simple scene} group for all the scenes with $4$ and $6$ blocks and a {\it complex scene} for the rest of scenes with $10$ and $14$ blocks. We investigate in two-direction classification, shown in Figure~\ref{fig:cross_group}, namely:
\begin{enumerate}
\item
Train on simple scenes and predict on complex scenes: Train on 4 and 6 blocks and test on 10 and 14 blocks
\item
Train on complex scenes and predict on simple scenes: Train on 10 and 14 blocks and test on 4 and 6 blocks
\end{enumerate}

The result is shown in Table~\ref{tab:cross_group}. When trained on simple scenes and predicting on complex scenes, it gets $69.9\%$, which is significantly better than random guess at $50\%$. This is understandable as the learned visual feature can transfer across different scene. Further we observe significant performance boost when trained on complex scenes and tested on simple scene. This can be explained by the richer feature learned from the complex scenes with better generalization. 

\begin{table*}\setlength{\tabcolsep}{7pt}
\centering
\ra{1.4}
\begin{tabular*}{0.75\textwidth}{l c c}\toprule
Setting & Simple $\rightarrow$ Complex & Complex $\rightarrow$ Simple\\
\hline
Accuracy (\%)& 69.9 & 86.9\\
\bottomrule
\end{tabular*}
\caption{Cross-group experiment results.}
\label{tab:cross_group}
\end{table*}

%--------------------------------------------------------------------------------------------------------------------------------------
\subsubsection{Generalization Experiment}
In this set of experiment, we want to explore if we can train a general model to predict stability for scenes with any scene parameters, which is very similar to human's prediction in the task. We use training images from all different scene groups and test on any groups. The Result is shown in Table~\ref{tab:general}. While the performance exhibits similar trend to the one in the intra-group with respect to the complexity of the scenes, namely increasing recognition rate for simpler settings and decreasing rate for more complex settings, there is a consistent improvement over the intra-group experiment for individual groups. Together with the result in the cross-group experiment, it suggests  a strong generalization capability of the image trained model.

\begin{table*}\setlength{\tabcolsep}{7pt}
\centering
\ra{1.4}
\begin{tabular*}{0.62\textwidth}{@{ }cccccc@{ }}\toprule
Num.of Blks & \multicolumn{2}{c}{Uni.} & \phantom{ab}& \multicolumn{2}{c}{NonUni.}\\
\cmidrule{2-3} \cmidrule{5-6}
 & $2D$ & $3D$ && $2D$ & $3D$\\ 
 \midrule
4B  &93.2 &99.0 &&95.4 &99.8 \\
6B  &89.0 &94.8 &&87.8 &93.0 \\
10B &83.4 &76.0 &&77.2 &74.8 \\
14B &82.4 &67.2 &&78.4 &66.2 \\ 
\bottomrule
\end{tabular*}
\caption{Results for generalization experiments.} % from prediction files
\label{tab:general}
\end{table*}

\subsubsection{Discussion}
Overall, we can conclude that direct stability prediction is possible and in fact fairly accurate at recognition rates over $80\%$ for moderate difficulty levels. As expected, the 3D setting adds difficulties to the prediction from appearance due to significant occlusion for towers of more than 10 blocks. Surprisingly, little effect was observed for small tower sizes switching from uniform to non-uniform blocks - although the appearance difference can be quite small.

%--------------------------------------------------------------------------------------------------------------------------------------

\subsection{Human Judgment on Synthetic Data}
For human subject test, the predictions are binarized, namely ``definitely unstable'' and ``probably unstable'' are treated as unstable prediction and ``probably stable'' and ``definitely stable'' as stable prediction regardless of the certainty quantifiers. The results are shown in Table~\ref{tab:human_test} and Figure~\ref{fig:human_cnn}.

\subsubsection{How well do humans perform on this task?}
For very simple scenes with few blocks, human can reach close to perfect performance while for complex scenes, the performance drops significantly to around $60\%$. 

\subsubsection{How does human performance vary with respect to scene variations}
As we discuss before, the number of blocks indicates the scene's complexity, given the same block size and stacking depth condition, the human's prediction degrades with increasing number of blocks in the scene in general. Given the same number block size condition and the number of blocks in the scene, the human's predictions in 3D stacking are better than the counterpart in 2D. This can partially be explained by the factor that the structure can have larger chance to be lower than the scene with stacking constraints, and the decreased height in return reduces the scene's complexity for human's judgment. The varied blocks size consistently shows higher difficulty than the fixed blocks size as in most cases, when one scene group changes the block condition from ``Uni'' to ``NonUni'', the performance decreases.

\subsubsection{How does human performance compare to image-based model?}
Compared to the human prediction in the same part of test data, the image-based model outperforms human in most scene groups. While showing similar trends in performance with respect to different scene parameters, the image-based model is less affected by a more difficult scene parameter setting, for example, given the same block size and stacking depth condition, the prediction accuracy decreases more slowly than the counter part in human prediction. We interpret this as image-based model possesses better generalization capability than human in the very task.  

\begin{table*}\setlength{\tabcolsep}{7pt}
\centering
\ra{1.4}
\begin{tabular*}{0.95\textwidth}{@{ }cccccc@{ }}\toprule
Num.of Blks & \multicolumn{2}{c}{Uni.} & \phantom{ab}& \multicolumn{2}{c}{NonUni.}\\
\cmidrule{2-3} \cmidrule{5-6}
& $2D$ & $3D$ && $2D$ & $3D$\\ 
 \midrule
4B  &79.1/{\bf91.7} &93.8/{\bf 100.0} &&72.9/{\bf 93.8} &92.7/{\bf 100.0} \\
6B  &78.1/{\bf 91.7} &83.3/{\bf 93.8} &&71.9/{\bf 87.5} &89.6/{\bf 93.8} \\
10B &67.7/{\bf 87.5} &72.9/72.9 &&66.7/{\bf 72.9} &71.9/68.8 \\
14B &71.9/{\bf 79.2} &68.8/66.7 &&{\bf 71.9}/{\bf 81.3} &59.3/{\bf 60.4} \\ 
\bottomrule
\end{tabular*}
\caption{Results from human subject test $a$ and corresponded accuracies from image-based model $b$ in format $a/b$ for the sampled data.} % from prediction files
\label{tab:human_test}
\end{table*}

\begin{figure}
\centering
        \begin{subfigure}[b]{0.4\textwidth}
                \centering
                \includegraphics[width=.85\linewidth]{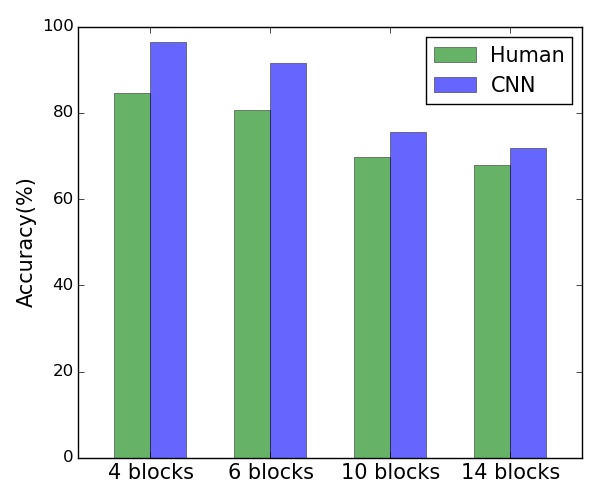}
                \caption{Blocks number}
                \label{fig:blks_alone}
        \end{subfigure}%
        \begin{subfigure}[b]{0.26\textwidth}
                \centering
                \includegraphics[width=.85\linewidth]{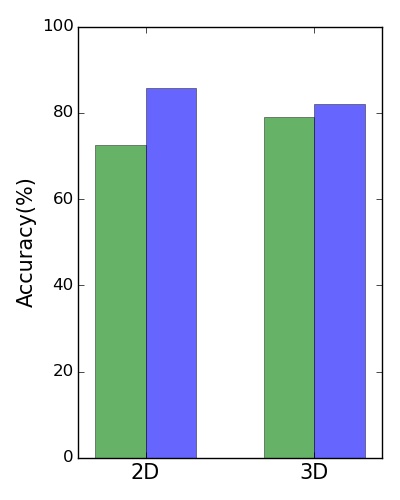}
                \caption{Stacking depth}
                \label{fig:dims_alone}
        \end{subfigure}%
        \begin{subfigure}[b]{0.26\textwidth}
                \centering
                \includegraphics[width=.85\linewidth]{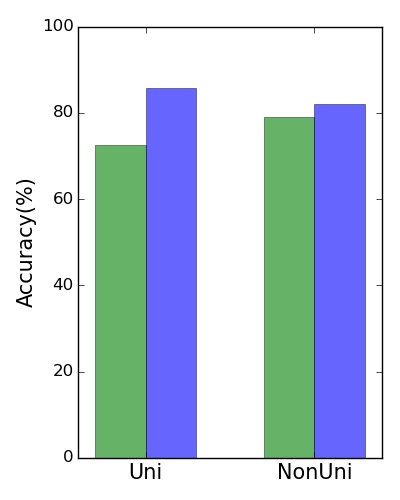}
                \caption{Blocks sizes}
                \label{fig:szs_alone}
        \end{subfigure}%
        \caption{Average accuracies for each scene parameters.}
        \label{fig:parameters_alone}
\end{figure}

To gain further insight into the results, we plot the average accuracy against each scene parameter alone. The results are shown in Figure~\ref{fig:parameters_alone}. Both human and the image-based model decrease consistently with respect to the number of blocks in the scene. However for both stacking depth and block size, human and the image-based model exhibit different trends, while the image-based model always outperforms the human, the human performance catches up in more complex scene parameter settings.

\subsubsection{How do failure cases differ? Are they plausible?}
\begin{figure}
\begin{center}$
%\addtolength{\tabcolsep}{-4.5pt}
\begin{tabular}{cccc}
\label{fig:fhum1}\includegraphics[width=0.22\linewidth]{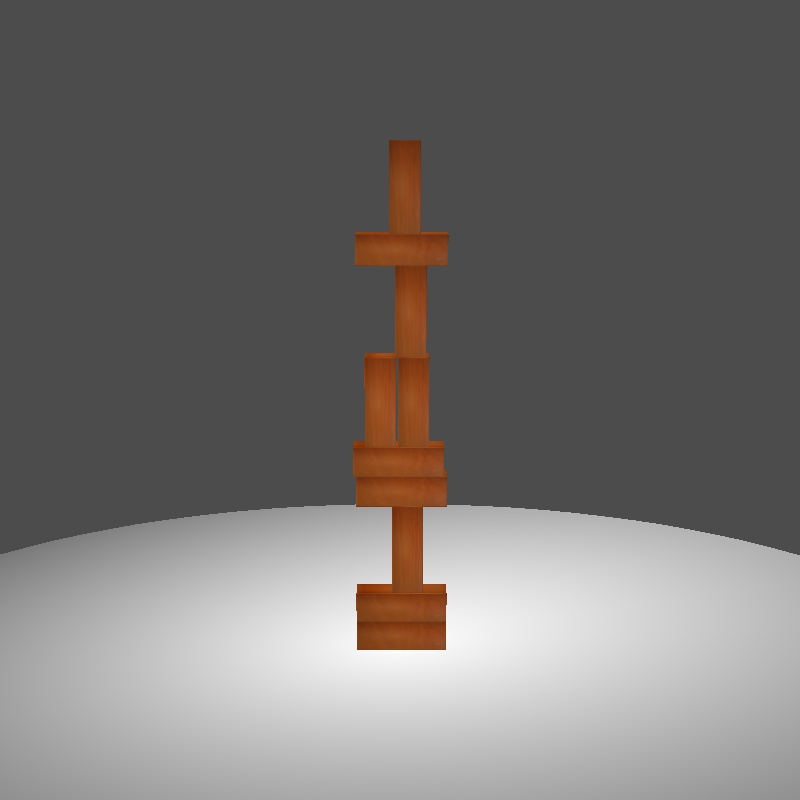}&
\label{fig:fhum2}\includegraphics[width=0.22\linewidth]{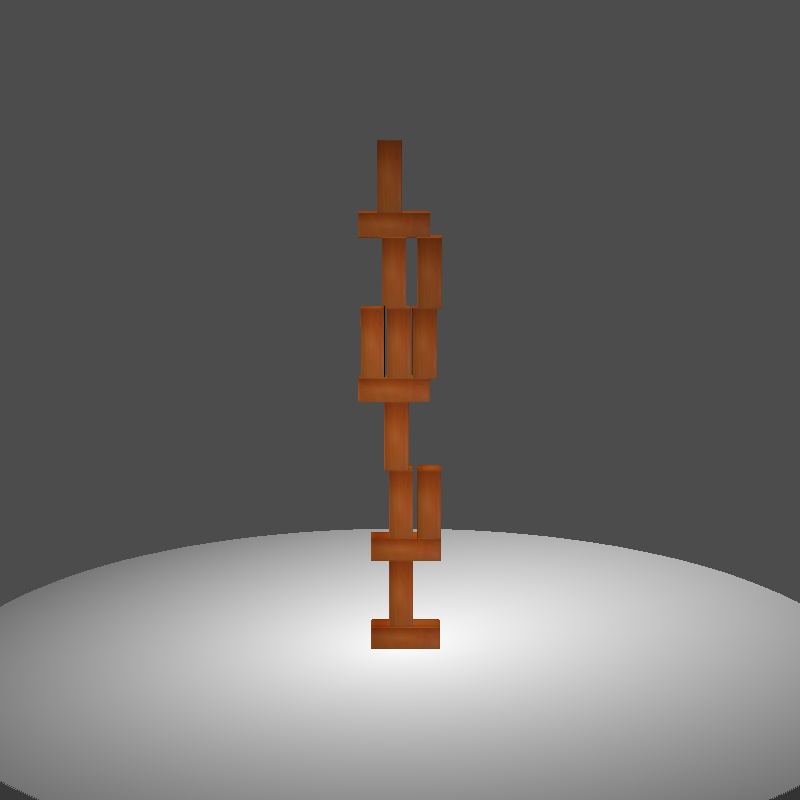}&
\label{fig:fhum3}\includegraphics[width=0.22\linewidth]{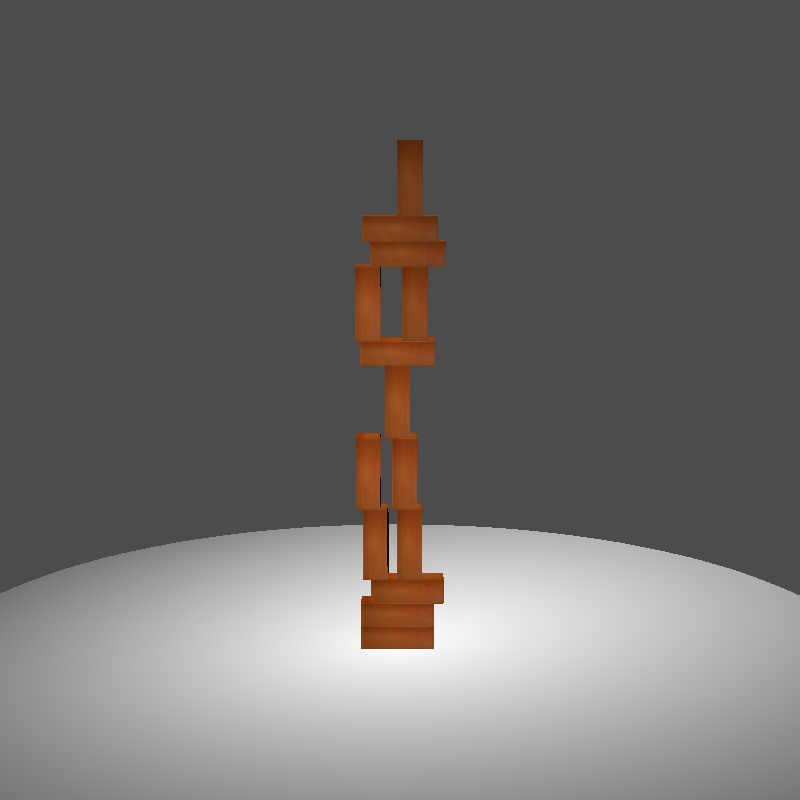}&
\label{fig:fhum4}\includegraphics[width=0.22\linewidth]{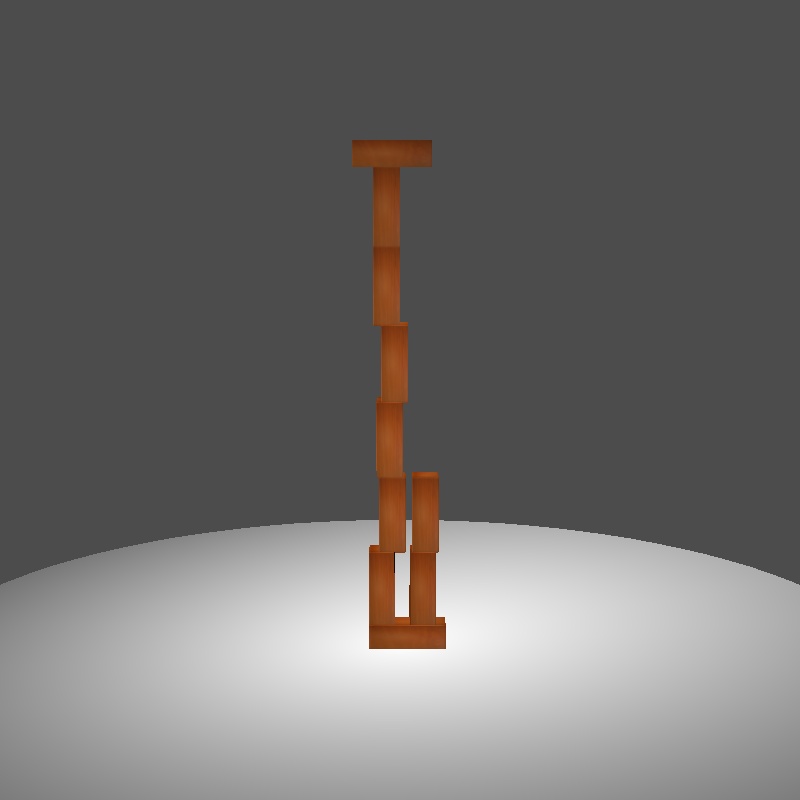}\\
\label{fig:fhum1}\includegraphics[width=0.22\linewidth]{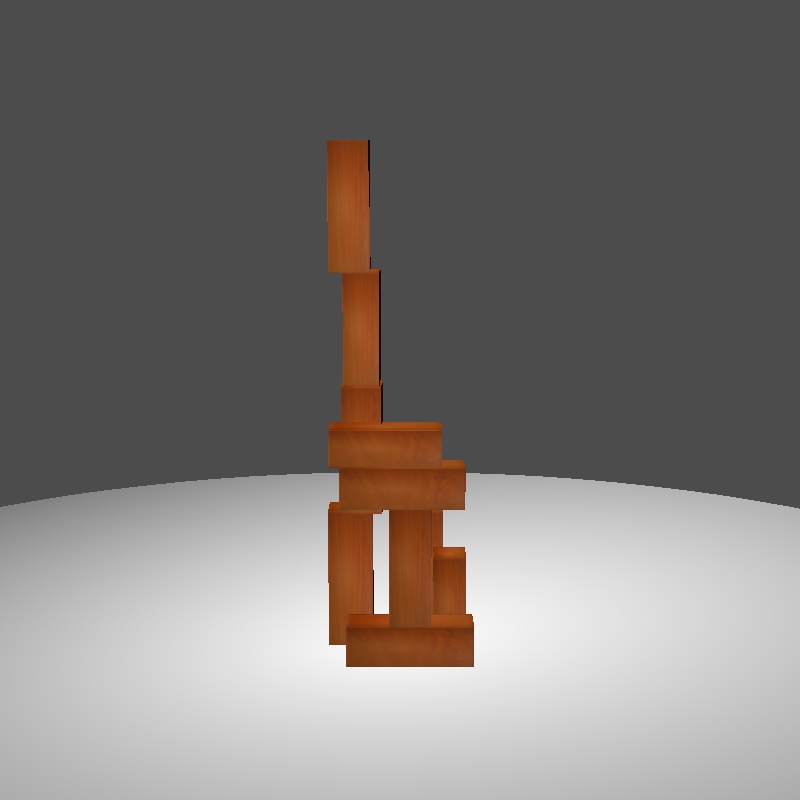}&
\label{fig:fhum2}\includegraphics[width=0.22\linewidth]{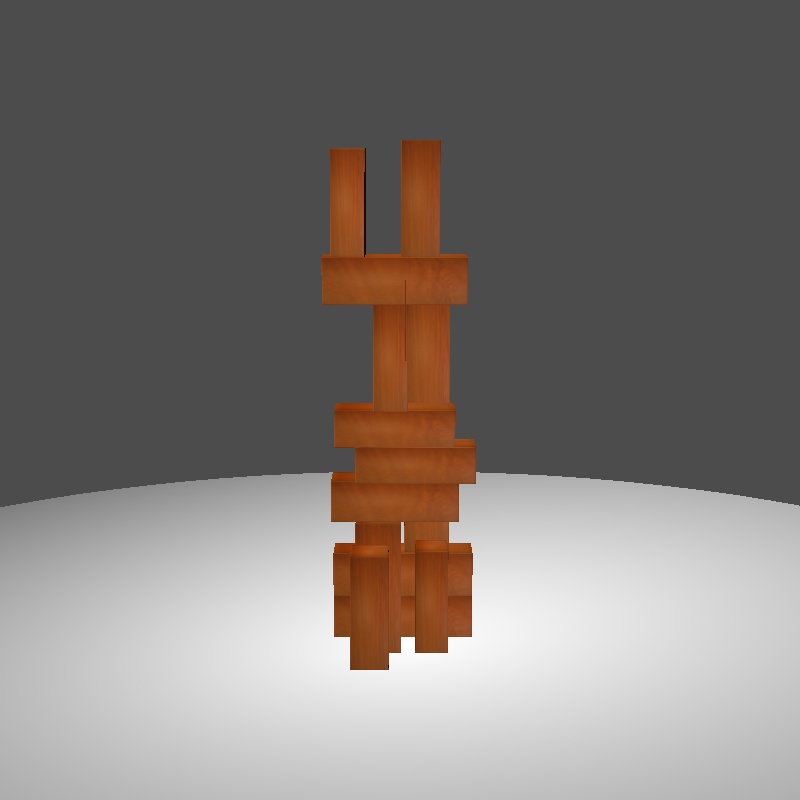}&
\label{fig:fhum3}\includegraphics[width=0.22\linewidth]{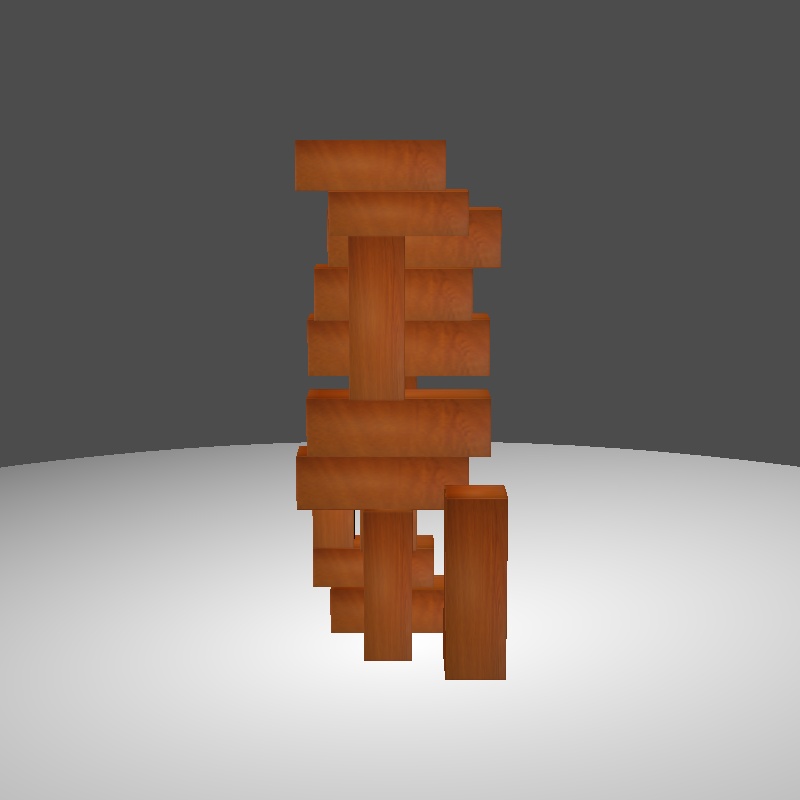}&
\label{fig:fhum4}\includegraphics[width=0.22\linewidth]{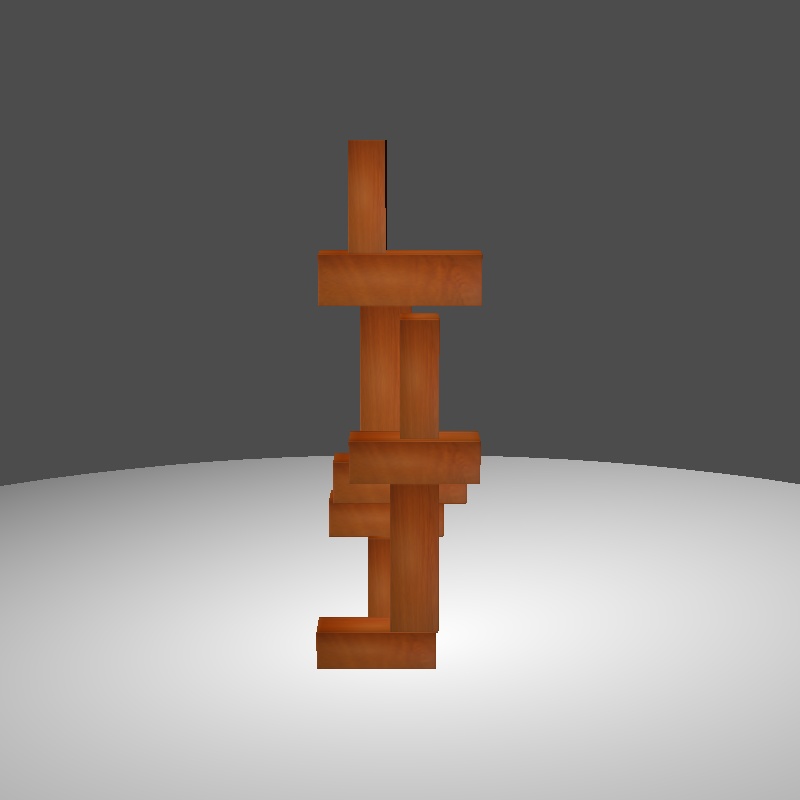}\\
\end{tabular}$
\end{center}
\caption{Top row, example of failure predictions as unstable from human where the machine made correct decisions; bottom row, example of failure predictions as unstable from machine where human made it right.}
\label{fig:failure}
\end{figure}

\begin{figure}[h]
\begin{center}
\includegraphics[width=0.7\linewidth]{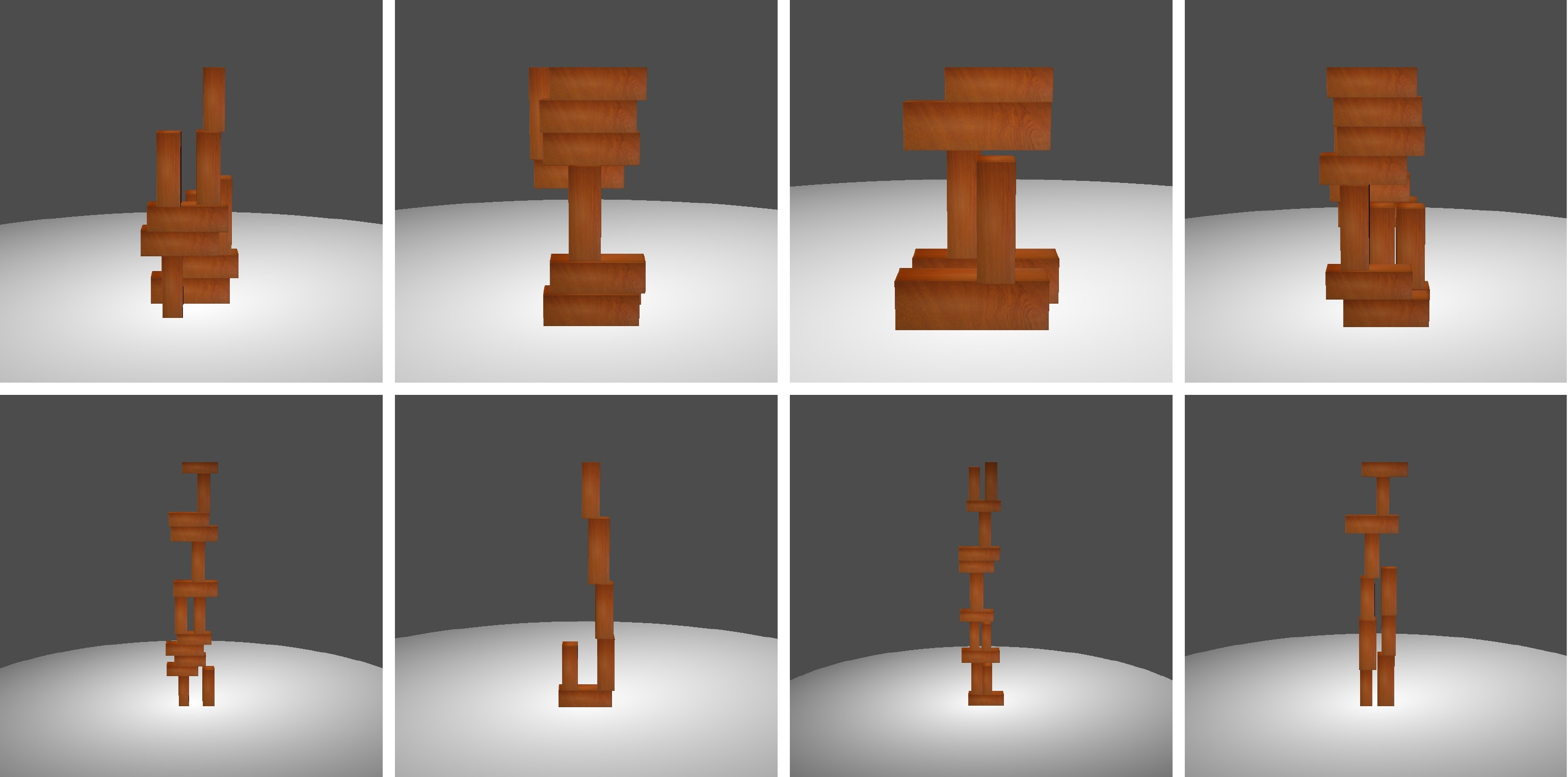}
\end{center}
\caption{Top row, example of failure predictions as stable from human where the machine made correct decisions; bottom row, example of failure predictions as stable from machine where human made it right.}
\label{fig:failure_more}
\end{figure}

In our test, it shows that human prone to make mistake for scenes in significant height while the machine is less affected by the factor. This is also consistent with the observation in \cite{battaglia2013simulation} that height plays an important role in human's judgment for stability. In contrast, the machine is trained across different heights, and hence  can adapt to more variation. Top row in Figure~\ref{fig:failure} shows some examples of such scenes. On the other hand, the machine makes more mistake when the scenes are constructed multiple layers than human. This is understandable as our model is only trained on monocular images while the human has the prior knowledge for perception of depth information. Examples are shown in the bottom row in Figure~\ref{fig:failure}. 
Figure~\ref{fig:failure_more} provides further examples of some false predictions of stable from unstable for both human and machine. While the occlusion condition can both affect human and machine for the judgment, it affects the machine more than human which is consistent with the false prediction of unstable. Similarly, height affects human more than machine as in false unstable predictions.

\begin{figure}
\centering
%\addtolength{\tabcolsep}{-4.5pt}
\begin{tabular}{cc}
\label{fig:human_rate}\includegraphics[width=0.4\linewidth]{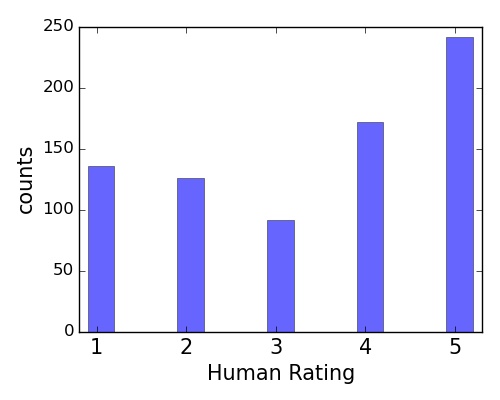}&
\label{fig:cnn_rate}\includegraphics[width=0.4\linewidth]{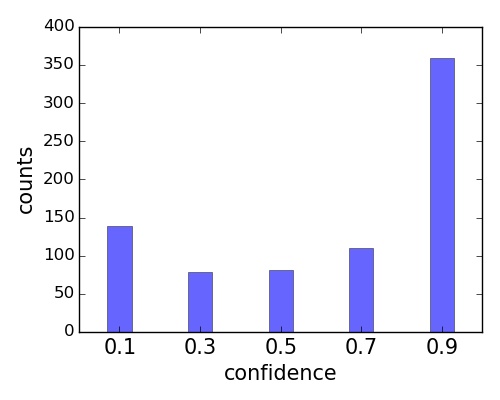}\\
\end{tabular}
\caption{Human rating distribution (left) and confidence distribution from CNN prediction (right).}
\label{fig:all_hist}
\end{figure}

Further, we count the histogram of human's rating and the prediction confidence from the image based model (for visualization purpose, we quantized the prediction confidence into 5 bins). The result is shown in Figure~\ref{fig:all_hist}. It's interesting to see the two distributions are relatively similar.

\begin{figure}
\centering
%\addtolength{\tabcolsep}{-4.5pt}
\begin{tabular}{cc}
\label{fig:4b_scatter}\includegraphics[width=0.37\linewidth]{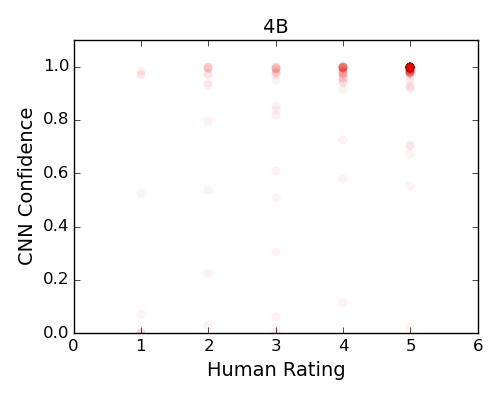}&
\label{fig:6b_scatter}\includegraphics[width=0.37\linewidth]{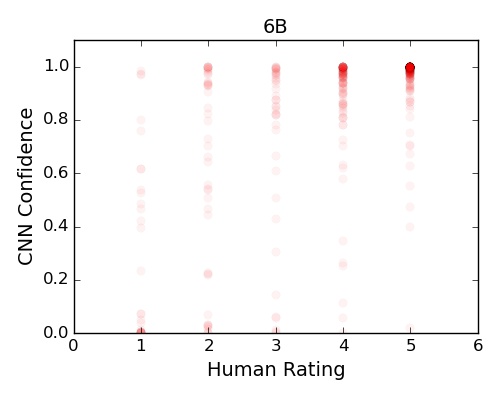}\\
\label{fig:10b_scatter}\includegraphics[width=0.37\linewidth]{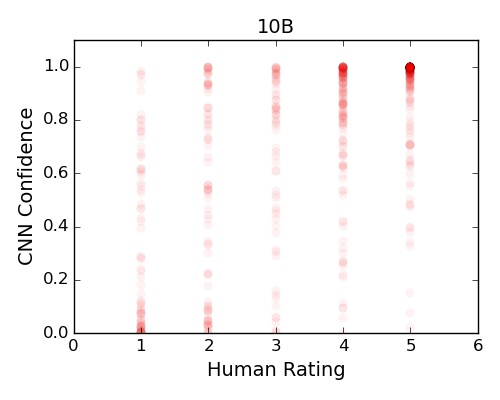}&
\label{fig:14b_scatter}\includegraphics[width=0.37\linewidth]{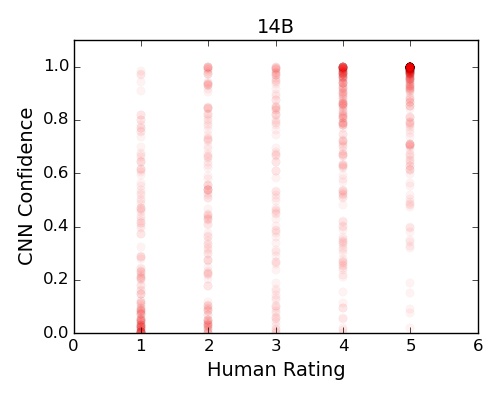}\\
\end{tabular}
\caption{Predictions of CNN model against human prediction for different number of blocks.} 
\label{fig:human_cnn_blks}
\end{figure}

\begin{figure}
\centering
%\addtolength{\tabcolsep}{-4.5pt}
\begin{tabular}{cc}
\label{fig:2d_scatter}\includegraphics[width=0.37\linewidth]{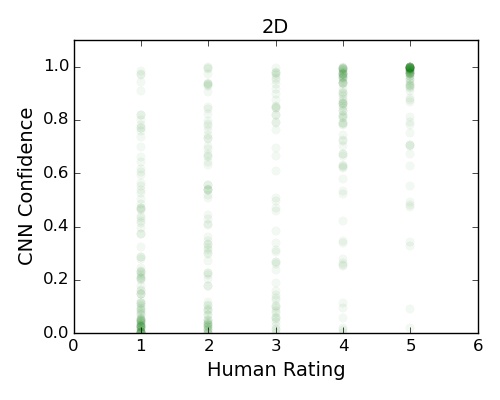}&
\label{fig:3d_scatter}\includegraphics[width=0.37\linewidth]{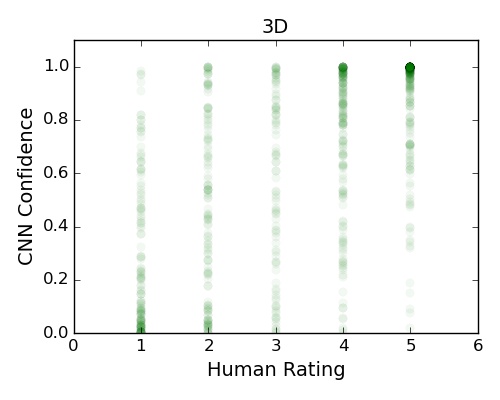}\\
\end{tabular}
\caption{Predictions of CNN model against human prediction for different stacking depths. }
\label{fig:human_cnn_depth}
\end{figure}

\begin{figure}
\centering
%\addtolength{\tabcolsep}{-4.5pt}
\begin{tabular}{cc}
\label{fig:uni_scatter}\includegraphics[width=0.37\linewidth]{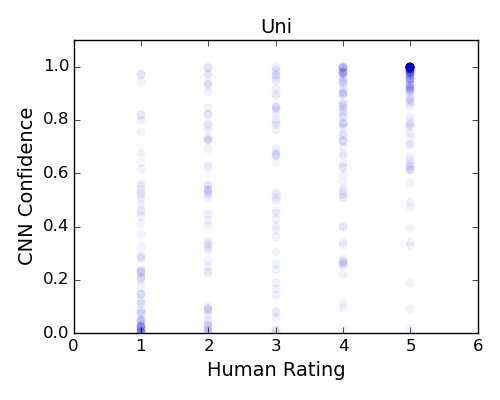}&
\label{fig:nonuni_scatter}\includegraphics[width=0.37\linewidth]{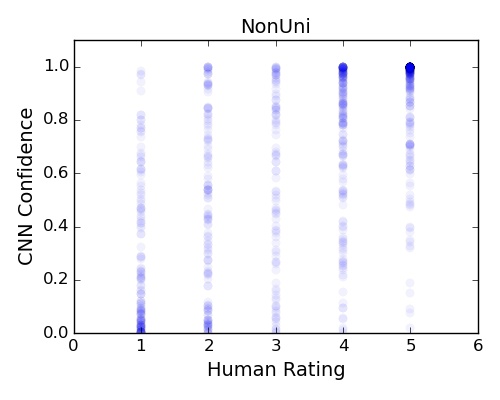}\\
\end{tabular}
\caption{Predictions of CNN model against human prediction for different block size conditions. }
\label{fig:human_cnn_size}
\end{figure}

\paragraph{{\bf Correlation between human performance and machine}}
Different from \cite{battaglia2013simulation}, our work does not aim to reconstruct human's inner mechanism hence the correlation between the human's prediction and the model's is not our priority. Yet we list such statistics to provide a more comprehensive image of the results to the reader. Here we shown the scatter plots for the pair $(x,y)$ of human's prediction $x$ and machine's prediction $y$ by different scene parameters, namely number of blocks (Figure~\ref{fig:human_cnn_blks}), stacking depth (Figure~\ref{fig:human_cnn_depth}) and block sizes (Figure~\ref{fig:human_cnn_size}). We computed the Pearson correlation coefficient for each group. The detailed values are shown in Table~\ref{tab:corr_blks}~\ref{tab:corr_depth}~\ref{tab:corr_size}. Interestingly, human prediction and human prediction are moderately positive correlated.

%%%%%%%%%%% moved conclusion here
\section{Conclusion}
In this work, we answer to the question if and how well we can build up a mechanism to predict physical stability directly from visual input. In contrast to existing approaches we bypass explicit 3D representation and physical simulation and learn a model for visual stability prediction from data. We evaluate our model on a range of conditions including variations in number of blocks, size of blocks and 3D structure of the overall tower. The results reflect the challenges of an increasing complex inference with increasing size of the structure as well as challenges due to small features the stability hinges on due to occlusions or block size variations.

Based on these encouraging results we envision systems that exploit such data driven notions of physics to arrive at advanced methods for scene understanding that reason on physical plausible state during visual inference. We also will investigate richer output spaces than binary labels that shed more light on the quality of physical understanding that was acquired by the learning based approach.

\begin{figure}
\begin{center}$
\addtolength{\tabcolsep}{-4.pt}
\begin{tabular}{cc}
\label{fig:2d_uni}\includegraphics[width=0.37\linewidth]{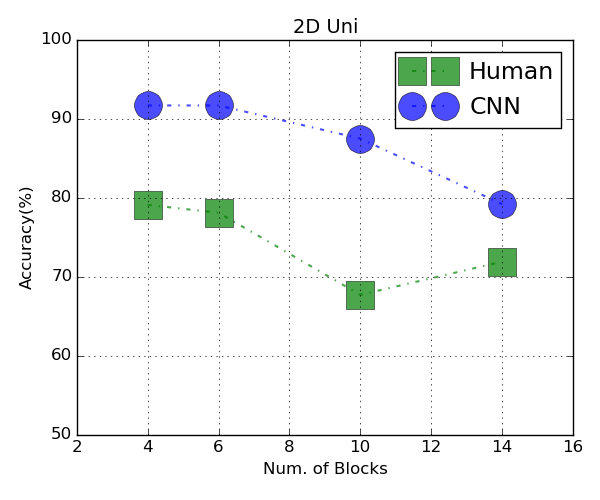}&
\label{fig:3d_uni}\includegraphics[width=0.37\linewidth]{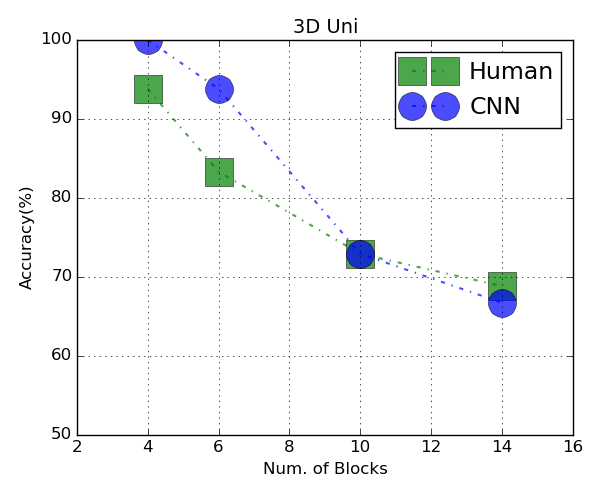}\\
\label{fig:2d_nonuni}\includegraphics[width=0.37\linewidth]{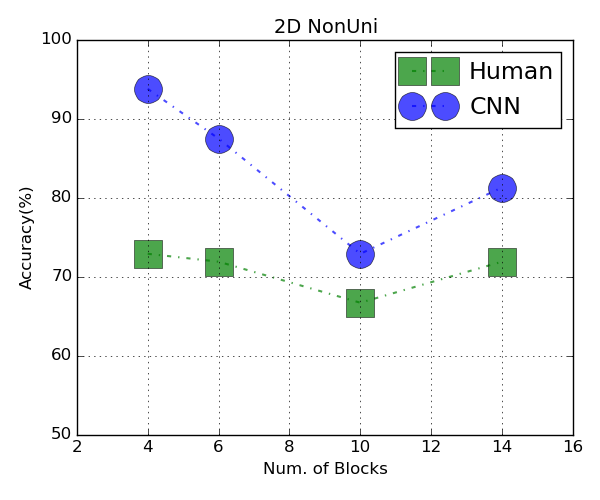}&
\label{fig:3d_nonuni}\includegraphics[width=0.37\linewidth]{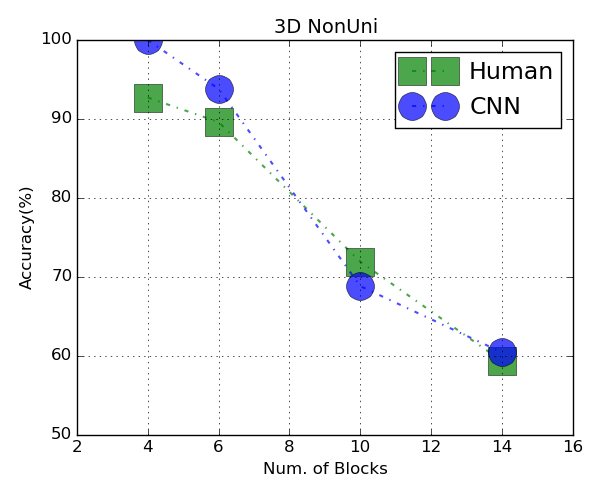}\\
\end{tabular}$
\end{center}
\caption{Overview of comparison results between human subject test and image-based model.}
\label{fig:human_cnn}
\end{figure}

\begin{table}\setlength{\tabcolsep}{7pt}
\centering
\ra{1.4}
\begin{tabular*}{0.6\textwidth}{l c c c c}\toprule
Setting & 4B & 6B & 10B & 14B\\
\hline
Corr. (\%)& 0.479 & 0.656 & 0.437 & 0.485\\
\bottomrule
\end{tabular*}
\caption{Correlation coefficients against number of blocks.}
\label{tab:corr_blks}
\end{table}

\begin{table}\setlength{\tabcolsep}{7pt}
\centering
\ra{1.4}
\begin{tabular*}{0.4\textwidth}{l c c}\toprule
Setting & 2D & 3D\\
\hline
Corr. (\%)& 0.596 & 0.575\\
\bottomrule
\end{tabular*}
\caption{Correlation coefficients against stacking depth.}
\label{tab:corr_depth}
\end{table}

\begin{table}\setlength{\tabcolsep}{7pt}
\centering
\ra{1.4}
\begin{tabular*}{0.4\textwidth}{l c c c c}\toprule
Setting & Uni & NonUni\\
\hline
Corr. (\%)& 0.667 & 0.587\\
\bottomrule
\end{tabular*}
\caption{Correlation coefficients against different block size conditions.}
\label{tab:corr_size}
\end{table}

%\par\vfill\par

\clearpage

\bibliographystyle{splncs03}
\bibliography{mybib}

\begin{thebibliography}{10}
\providecommand{\url}[1]{\texttt{#1}}
\providecommand{\urlprefix}{URL }

\bibitem{baillargeon1995model}
Baillargeon, R.: A model of physical reasoning in infancy. Advances in infancy
  research  (1995)

\bibitem{baillargeon1994infants}
Baillargeon, R.: How do infants learn about the physical world? Current
  Directions in Psychological Science  (1994)

\bibitem{baillargeon2002acquisition}
Baillargeon, R.: The acquisition of physical knowledge in infancy: A summary in
  eight lessons. Blackwell handbook of childhood cognitive development  (2002)

\bibitem{baillargeon2008innate}
Baillargeon, R.: Innate ideas revisited: For a principle of persistence in
  infants' physical reasoning. Perspectives on Psychological Science  (2008)

\bibitem{battaglia2013simulation}
Battaglia, P.W., Hamrick, J.B., Tenenbaum, J.B.: Simulation as an engine of
  physical scene understanding. Proceedings of the National Academy of Sciences
   (2013)

\bibitem{cholewiak2013visual}
Cholewiak, S.A., Fleming, R.W., Singh, M.: Visual perception of the physical
  stability of asymmetric three-dimensional objects. Journal of vision  (2013)

\bibitem{cholewiak2015perception}
Cholewiak, S.A., Fleming, R.W., Singh, M.: Perception of physical stability and
  center of mass of 3-d objects. Journal of vision  (2015)

\bibitem{coumans2010bullet}
Coumans, E.: Bullet physics engine. Open Source Software: http://bulletphysics.
  org  1 (2010)

\bibitem{deng2009imagenet}
Deng, J., Dong, W., Socher, R., Li, L.J., Li, K., Fei-Fei, L.: Imagenet: A
  large-scale hierarchical image database. In: CVPR (2009)

\bibitem{fragkiadaki2015learning}
Fragkiadaki, K., Agrawal, P., Levine, S., Malik, J.: Learning visual predictive
  models of physics for playing billiards. arXiv preprint arXiv:1511.07404
  (2015)

\bibitem{goslin2004panda3d}
Goslin, M., Mine, M.R.: The panda3d graphics engine. Computer  (2004)

\bibitem{gupta2010blocks}
Gupta, A., Efros, A.A., Hebert, M.: Blocks world revisited: Image understanding
  using qualitative geometry and mechanics. In: ECCV (2010)

\bibitem{hamrick2011internal}
Hamrick, J., Battaglia, P., Tenenbaum, J.B.: Internal physics models guide
  probabilistic judgments about object dynamics. In: Proceedings of the 33rd
  annual conference of the cognitive science society. Cognitive Science Society
  Austin, TX (2011)

\bibitem{jia2014caffe}
Jia, Y., Shelhamer, E., Donahue, J., Karayev, S., Long, J., Girshick, R.,
  Guadarrama, S., Darrell, T.: Caffe: Convolutional architecture for fast
  feature embedding. In: Proceedings of the ACM International Conference on
  Multimedia. ACM (2014)

\bibitem{krizhevsky2012imagenet}
Krizhevsky, A., Sutskever, I., Hinton, G.E.: Imagenet classification with deep
  convolutional neural networks. In: NIPS (2012)

\bibitem{lecun1995comparison}
LeCun, Y., Jackel, L., Bottou, L., Brunot, A., Cortes, C., Denker, J., Drucker,
  H., Guyon, I., Muller, U., Sackinger, E., et~al.: Comparison of learning
  algorithms for handwritten digit recognition. In: International conference on
  artificial neural networks (1995)

\bibitem{fergus16blocsarxiv}
Lerer, A., Gross, S., Fergus, R.: Learning physical intuition of block towers
  by example. arXiv preprint arXiv:1603.01312  (2016)

\bibitem{li12eccv}
Li, W., Fritz, M.: Recognizing materials from virtual examples. In: ECCV (2012)

\bibitem{macdougal2012galileo}
MacDougal, D.W.: Galileo’s great discovery: How things fall. In: Newton's
  Gravity. Springer (2012)

\bibitem{mccloskey1983intuitive}
McCloskey, M.: Intuitive physics. Scientific american  (1983)

\bibitem{mottaghi2015newtonian}
Mottaghi, R., Bagherinezhad, H., Rastegari, M., Farhadi, A.: Newtonian image
  understanding: Unfolding the dynamics of objects in static images. arXiv
  preprint arXiv:1511.04048  (2015)

\bibitem{peng2015learning}
Peng, X., Sun, B., Ali, K., Saenko, K.: Learning deep object detectors from 3d
  models. In: ICCV (2015)

\bibitem{razavian2014cnn}
Razavian, A., Azizpour, H., Sullivan, J., Carlsson, S.: Cnn features
  off-the-shelf: an astounding baseline for recognition. In: CVPR workshops
  (2014)

\bibitem{rematas16cvpr}
Rematas, K., Ritschel, T., Fritz, M., Gavves, E., Tuytelaars, T.: Deep
  reflectance maps. In: CVPR (2016)

\bibitem{kostas14cvpr}
Rematas, K., Ritschel, T., Fritz, M., Tuytelaars, T.: Image-based synthesis and
  re-synthesis of viewpoints guided by 3d models. In: CVPR (2014)

\bibitem{silberman2012indoor}
Silberman, N., Hoiem, D., Kohli, P., Fergus, R.: Indoor segmentation and
  support inference from rgbd images. In: ECCV (2012)

\bibitem{simonyan2014very}
Simonyan, K., Zisserman, A.: Very deep convolutional networks for large-scale
  image recognition. arXiv preprint arXiv:1409.1556  (2014)

\bibitem{Smith1994}
Smith, B., Casati, R.: Naive Physics: An Essay in Ontology. Philosophical
  Psychology (1994)

\bibitem{wu2015galileo}
Wu, J., Yildirim, I., Lim, J.J., Freeman, B., Tenenbaum, J.: Galileo:
  Perceiving physical object properties by integrating a physics engine with
  deep learning. In: NIPS (2015)

\bibitem{Xie_2013_ICCV}
Xie, D., Todorovic, S., Zhu, S.C.: Inferring "dark matter" and "dark energy"
  from videos. In: ICCV (2013)

\bibitem{zheng2013beyond}
Zheng, B., Zhao, Y., Yu, J., Ikeuchi, K., Zhu, S.C.: Beyond point clouds: Scene
  understanding by reasoning geometry and physics. In: CVPR (2013)

\end{thebibliography}
\end{document}